%% file: main.tex
\documentclass[letterpaper, 10 pt, conference]{ieeeconf}

\IEEEoverridecommandlockouts                              

\usepackage{graphicx} 
\usepackage{epsfig} 
\usepackage{dblfloatfix} 
\usepackage{courier}
\usepackage{amsmath} 
\usepackage{amssymb}  
\usepackage{xcolor}
\usepackage{amsmath} 
\usepackage{amssymb}
\usepackage{float}
\usepackage{algorithm}		
\usepackage{algorithmicx} 	
\usepackage{algpseudocode}  
\usepackage{todonotes}
\usepackage{multicol}
\usepackage[backend=bibtex,bibstyle=ieee,citestyle=numeric-comp,doi=false,isbn=false,eprint=false]{biblatex}
\title{\LARGE \bf
Optimal Multi-robot Formations for Relative Pose Estimation Using Range Measurements
}

\author{Charles Champagne Cossette, Mohammed Ayman Shalaby,  David Saussi\'e, J\'er\^ome Le Ny, James Richard Forbes
\thanks{*This work was supported by the FRQNT under grant 2018-PR-253646, with funding also acknowledged from CFI JELF, the William Dawson Scholar Program, and the NSERC Discovery Grant Program.}
\thanks{C. C. Cossette, M. A. Shalaby, J. R. Forbes are with the Department of Mech. Engineering, McGill University. {\tt \small charles.cossette@mail.mcgill.ca, mohammed.shalaby@mail.mcgill.ca, james.richard.forbes@mcgill.ca}}%
\thanks{D. Saussi\'e, J. Le Ny, are with the Department of Electrical Engineering, Polytechnique Montr\'eal. 
        {\tt\small d.saussie@polymtl.ca} {\tt\small jerome.le-ny@polymtl.ca}}%
}

\input{Commands.tex}

\definecolor{darkgreen}{rgb}{0,0.5,0}

\begin{document}

\maketitle
\thispagestyle{empty}
\pagestyle{empty}

\begin{abstract}
    In multi-robot missions, relative position and attitude information between agents is valuable for a variety of tasks such as mapping, planning, and formation control. In this paper, the problem of estimating relative poses from a set of inter-agent range measurements is investigated. Specifically, it is shown that the estimation   accuracy is highly dependent on the true relative poses themselves, which prompts the desire to find multi-agent formations that provide the best estimation performance. By direct maximization of Fischer information, it is shown in simulation and experiment that large improvements in estimation accuracy can be obtained by optimizing the formation geometry of a team of robots.    
\end{abstract}

\section{Introduction}
The ability for a robot, or \emph{agent}, to determine the relative position and attitude, collectively called \emph{pose}, of another robot is an important prerequisite in multi-robot team applications. Tasks such as collaborative mapping and planning, as well as formation control, usually require relative position or pose information between the robots. This functionality has been achieved using various sensors, such as cameras with object detection \cite{Li2021}, or with infrared emitters/receivers \cite{Mao2013}.

Ultra-wideband (UWB) is a type of radio signal that can be timestamped with sub-nanosecond-level accuracy at both transmission and reception \cite{Sahinoglu2008a}. As such, UWB is commonly used to obtain about 10-cm-accurate range (distance) measurements between a pair of UWB transceivers called \emph{tags}. The transceivers' small size, weight, and cost make them an attractive sensor for many robotics applications, including relative position estimation in multi-robot scenarios. By placing one or more tags on each robotic agent, a completely self-contained relative positioning solution is possible \cite{Guler2018,Nguyen2018}, which does not depend on any external infrastructure such as static UWB tags, called \emph{anchors}, or a motion capture system.

In this theme of infrastructure-free relative position estimation, a wide variety of approaches exist in the literature. For example, visual odometry or optical flow have been used along with a single UWB tag on each agent \cite{Guo2020, Nguyen2019, Nguyen2020,Cossette2021, Nguyen2022}. However, these single-tag-per-agent approaches typically have a \emph{persistency of excitation} (POE) requirement. That is, agents must be under persistent relative motion for relative states to be observable \cite{Batista2011,She2020}. This can be energy intensive and impractical, as a static or slowly-moving team of agents will have drifting position estimates. One way to eliminate the POE requirement is to use visual detection of other agents, as in \cite{Xu2020}, which also uses visual odometry and UWB ranging. Although their solution is accurate, deep-learning-based object detection can be computationally expensive, and the agents must periodically enter each other's camera field-of-view.

Another class of approaches that do not require computer vision or POE is to have multiple tags on some or all of the agents \cite{Guler2018,Hepp2016}. We have recently proposed installing two UWB tags on each agent \cite{Shalaby2021}, where we show that relative positions are observable from the range measurements alone. When combined with an inertial measurement unit (IMU) and a magnetometer, the agents' individual attitudes can be estimated relative to a world frame, allowing relative positions to be resolved in the world frame. However, magnetometer sensor measurements are substantially disturbed in the presence of metallic structures indoors \cite{Kok2018, Solin2018}, which degrades estimation accuracy. Another challenge is that there are certain formation geometries that cause the relative positions to be unobservable, such as when all the UWB tags lie on the same line \cite{Shalaby2021}. This is closely tied to the well-known general dependence of positioning accuracy on the geometry of the tags, and arises even in the presence of static UWB anchors \cite{Zhao2018a}. 

To avoid divergence of the state estimator, multi-robot missions relying on inter-robot range measurements for relative position estimation must avoid these aforementioned unobservable formation geometries. This imposes a constraint on planning algorithms. A planning solution to avoid unobservable positions is proposed in \cite{LeNy2018a}, where a cost function based on the Cram\'er-Rao bound quantifies the estimation accuracy as a function of robot positions. A similar approach is presented in \cite{Cano2021} for multi-tag robots. Limitations of these approaches include the requirement of the presence of anchors, as well as the lack of explicit consideration of agent attitudes.

This paper presents a method for computing optimal formations for relative pose estimation, and is the first to do so in the absence of anchors. Furthermore, it is shown that with two-tag agents, both the relative position and relative heading of the agents are locally observable from range measurements alone. The problem setup is deliberately formulated in the agents' body frames, thus being completely invariant to any arbitrary world frame, eliminating the need for a magnetometer. This paper further differs from \cite{Cano2021} by using $SE(n)$ pose transformation matrices to represent the relative poses, avoiding the complications associated with angle parameterizations of attitude. This leads to the use of an \emph{on-manifold gradient descent} procedure to determine optimal formations. Simulations and experiments show that the variance of estimation error does indeed decrease as the agents approach their optimal formations.

The proposed cost function is general to 2D or 3D translations, arbitrary measurement graphs, and any number of arbitrarily-located tags. Moreover, the proposed cost function goes to infinity when the agents approach unobservable configurations, meaning that its use naturally avoids such unobservable formation geometries. For these reasons, the cost function is amenable to a variety of future planning applications, such as to impose an inequality constraint on an indoor exploration planning problem.

The paper is outlined as follows. The problem setup, notation, and other preliminaries are described in Section \ref{sec:prelim}. The optimization setup and results are described in Section \ref{sec:optim}. The optimal formations are evaluated experimentally in Section \ref{sec:exp}.

\section{Problem Setup, Notation, and Preliminaries} \label{sec:prelim}
\begin{figure}[b]
    \centering
    \includegraphics[width =0.7\linewidth]{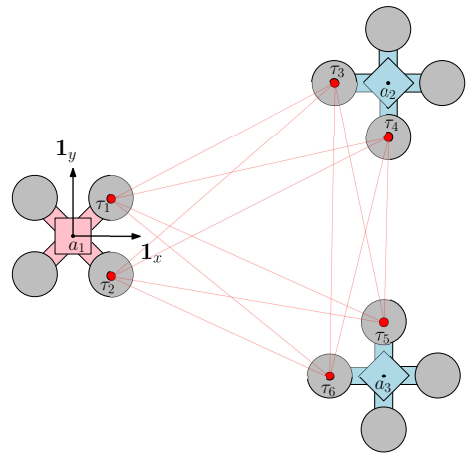}
    \caption{Problem setup and notation used. Each agent possesses a reference point $a_\alpha$ where $\alpha$ is the agent ID, as well as two tags $\tau_i, \tau_j$, where $i,j$ are the tag IDs. $\mbf{1}_x$ and $\mbf{1}_y$ are vectors which represent the $x$ and $y$ axis of Agent 1's body frame. Throughout this paper, the red agent is the arbitrary reference agent, and it will always be Agent 1 without loss of generality.}
    \label{fig:notation}
\end{figure}

Consider $N$ agents along with $M$ ranging tags distributed amongst them. Let $\tau_1, \tau_2, \ldots , \tau_M$ consist of unique physical points collocated with the ranging tags. Let $a_1, \ldots , a_N$ represent reference points on the agents themselves. The inter-tag range measurements are represented by a measurement graph $\mc{G} = (\mc{V}, \mc{E})$ where $\mc{V} = \{1, \ldots, M \}$ is the set of nodes, which is equivalent to the set of tag IDs, and $\mc{E}$ is the set of edges corresponding to the range measurements. Defining the set of agent IDs as $\mc{A} = \{1,\ldots, N \}$, it is convenient to define a simple ``lookup function'' $\ell : \mc{V} \to \mc{A}$ that returns the agent ID on which any particular tag is located. For example, if $\tau_i$ is physically on agent $\alpha$, then $\ell(i) = \alpha$. An example scenario with three agents using this notation is shown in Figure \ref{fig:notation}. A bolded $\mbf{1}$ and $\mbf{0}$ indicates an appropriately-sized identity and zero matrix, respectively.

\subsection{State Definition and Range Measurement Model}
Since the agents are rigid bodies, an orthonormal reference
frame attached to their bodies can be defined. A position vector representing the position of point $z$, relative to point $w$, resolved in the body frame of agent $\alpha$ is denoted $\mbf{r}^{zw}_\alpha \in \mathbb{R}^n$. The attitude of the body frame on agent $\alpha$ relative to the body frame on agent $\beta$ is represented with a rotation matrix $\mbf{C}_{\alpha \beta} \in SO(n)$ such that $\mbf{r}^{zw}_\alpha = \mbf{C}_{\alpha \beta} \mbf{r}^{zw}_\beta$. The relative position and attitude between agents $\alpha$ and $\beta$, $(\mbf{r}^{a_\beta a_\alpha}_\alpha, \mbf{C}_{\alpha \beta})$ define the relative pose between them, and can be packaged together in a pose transformation matrix 
\beq 
\mbf{T}_{\alpha \beta} = \bma{cc} \mbf{C}_{\alpha \beta} & \mbf{r}^{a_\beta a_\alpha}_\alpha \\ \mbf{0}& 1 \ema \in SE(n).
\eeq
The exponential and logarithmic maps of the special Euclidean group $SE(n)$ are denoted $\exp:\mathfrak{se}(n) \to SE(n)$ and $\ln:SE(n) \to \mathfrak{se}(n)$, respectively, where $\mathfrak{se}(n)$ is the Lie algebra of $SE(n)$. The common ``wedge'' operator $(\cdot)^\wedge:\mathbb{R}^m \to \mathfrak{se}(n)$ and ``vee'' operator $(\cdot)^\vee:\mathfrak{se}(n) \to \mathbb{R}^m$  are also used in this paper. For a more thorough background on matrix Lie groups, including expressions for the aforementionned operators, see \cite{Sola2018a,Barfoot2019}.

 Throughout this paper, Agent 1 will be the arbitary reference agent, such that the poses of all the other agents are expressed relative to Agent 1
\beq 
\mbf{x} = (\mbf{T}_{12}, \ldots, \mbf{T}_{1N}) .
\eeq 
A single generic range measurement between tag $i$ and tag $j$ is modelled as a function of the state $\mbf{x} = (\mbf{T}_{12}, \ldots, \mbf{T}_{1N}) $ with 
\begin{align}
y_{ij}(\mbf{x}) &= ||\mbf{C}_{1\alpha}\mbf{r}^{\tau_i a_{\alpha}}_{\alpha} + \mbf{r}^{a_\alpha a_1}_1 - (\mbf{C}_{1\beta}\mbf{r}^{\tau_j a_\beta}_\beta + \mbf{r}^{a_\beta a_1}_1)|| + v_{ij},
\end{align}
where $\alpha = \ell(i),$ $\beta = \ell(j)$, and $v_{ij} \sim \mc{N}(0, \sigma^2_{ij})$. This can be written compactly with the pose transformation matrices,
\begin{align}
   y_{ij}(\mbf{x}) &=  \norm{\mbf{D}\mbf{T}_{1\alpha} \bma{c}\mbf{r}^{\tau_i a_\alpha}_\alpha \\ 1 \ema - \mbf{D}\mbf{T}_{1\beta} \bma{c}\mbf{r}^{\tau_j a_\beta}_\beta \\ 1 \ema } + v_{ij}  \nonumber \\
    & \triangleq  \norm{\mbf{D}\mbf{T}_{1\alpha} \mbf{p}_i - \mbf{D}\mbf{T}_{1\beta} \mbf{p}_j } + v_{ij} \label{eq:meas_model}  ,  
\end{align}
where $\mbf{D} = [\mbf{1} \; \mbf{0}]$.  In fact, the state $\mbf{x} = (\mbf{T}_{12}, \ldots, \mbf{T}_{1N}) $, written here as a tuple of pose matrices, is an element of a Lie group of its own, 
\bdis 
\mbf{x} \in SE(n) \times \ldots \times SE(n) \triangleq SE(n)^{N-1}.
\edis 
The group operation for $SE(n)^{N-1}$ is the elementwise matrix multiplication of the pose matrices in two arbitrary tuples, and the group inverse is the elementwise matrix inversion of the elements of the tuple $\mbf{x}$. The $\oplus$ operator is defined here as
\beq 
\mbf{x} \oplus \delta \mbf{x} = \left(\mbf{T}_{12}\exp(\delta \mbs{\xi}_2^\wedge),\; \ldots\;, \mbf{T}_{1N}\exp(\delta \mbs{\xi}_N^\wedge)\right) ,
\eeq 
where $\delta\mbs{\xi}_i \in \mathbb{R}^m$, $\delta \mbf{x} = [\delta \mbs{\xi}_2^\trans\; \ldots \; \delta \mbs{\xi}_N^\trans]^\trans \in \mathbb{R}^{m(N-1)}$, and will be used throughout the paper.

\section{Optimization} \label{sec:optim}
The goal is to find the relative agent poses that, with respect to some metric, provide the best relative pose estimation results if the estimation were to be done exclusively using the range measurements. The metric chosen in this paper is based on Fischer information and the Cram\'er-Rao bound, which will be recalled here.
\begin{definition}[Fischer information matrix \cite{Bar-Shalom2001}] Let $\mbf{y} \in \mathbb{R}^q$ be a continuous random variable that is conditioned on a nonrandom variable $\mbf{x} \in \mathbb{R}^n$. The \emph{Fischer information matrix} (FIM) is defined as 
    \beq
        \mbf{I}(\mbf{x}) = \mathbb{E} \left[\left(\frac{\p \ln p(\mbf{y}|\mbf{x})}{\p \mbf{x}}\right)^\trans \left(\frac{\p \ln p(\mbf{y}|\mbf{x})}{\p \mbf{x}}\right) \right] \in \mathbb{R}^{n \times n},
    \eeq
where $\mathbb{E}[\cdot]$ is the expectation operator and $p(\cdot)$ denotes a probability density function.
\end{definition}

\begin{theorem}[Cram\'er-Rao Bound \cite{Bar-Shalom2001}] Let $\mbf{y} \in \mathbb{R}^q$ be a continuous random variable that is conditioned on $\mbf{x} \in \mathbb{R}^n$. Let $\mbfhat{x}(\mbf{y})$ be an unbiased estimator of $\mbf{x}$, i.e., $\mathbb{E}[\mbf{e}(\mbf{x})] = \mathbb{E}[\mbfhat{x}(\mbf{y})-\mbf{x}] = \mbf{0}$. The \emph{Cram\'er-Rao lower bound} states that
    \beq
    \mathbb{E}\left[\mbf{e}(\mbf{x}) \mbf{e}(\mbf{x})^\trans \right] \geq \mbf{I}^{-1}(\mbf{x}).
    \eeq \label{thm:crlb}
\end{theorem}
\begin{theorem}[FIM for a Gaussian PDF] Consider the nonlinear measurement model with additive Gaussian noise,
    \beq
    \mbf{y} = \mbf{g}(\mbf{x}) + \mbf{v}, \qquad \mbf{v} \sim \mc{N}(\mbf{0}, \mbf{R}). \label{eq:fim_gaussian}
    \eeq
The Fischer information matrix is given by 
\beq
\mbf{I}(\mbf{x}) = \mbf{H}(\mbf{x})^\trans \mbf{R}^{-1} \mbf{H}(\mbf{x}),
\eeq
where $\mbf{H}(\mbf{x}) = \p \mbf{g}(\mbf{x})/\p \mbf{x}$.
\end{theorem}

The Cram\'er-Rao bound represents the minimum variance achievable by any unbiased estimator. Hence, motivated by Theorem \ref{thm:crlb}, an estimation cost function $J_{\mathrm{est}}$ is defined 
\beq 
J_{\mathrm{est}}(\mbf{x}) = - \ln \det \mbf{I}(\mbf{x}), \label{eq:j_est}
\eeq 
which will be minimized with the agent relative poses $\mbf{x}$ as the optimization variables. The logarithm of the determinant of $\mbf{I}(\mbf{x})$ is one option amongst many choices of matrix norms, such as the trace or Frobenius norm. We have found the chosen cost function to behave well in terms of numerical optimization and, most importantly, goes to infinity when the FIM becomes non-invertible. The state $\mbf{x}$ is locally observable from measurements $\mbf{y}$ if the measurement Jacobian $\mbf{H}(\mbf{x})$ is full column rank, which also makes the FIM full rank. As will be seen in Section \ref{sec:cost}, non-invertibility of the FIM also corresponds to formations that result in unobservable relative poses, which should be avoided.

To create a measurement model in the form of \eqref{eq:fim_gaussian}, the range measurements are all concatenated into a single vector 
\bdis
\mbf{y}(\mbf{x}) = \underbrace{[\ldots \\ y_{ij}(\mbf{x})  \\ \ldots]^\trans}_{\triangleq \mbf{g}(\mbf{x})} + \mbf{v}, \quad  \forall (i,j) \in \mc{E}, \; \mbf{v} \sim \mc{N}(\mbf{0}, \mbf{R}),
\edis
where $\mbf{R} = \mathrm{diag}(\ldots,\sigma^2_{ij},\ldots)$. It would be possible to directly descend the cost in \eqref{eq:j_est} with an optimization algorithm such as gradient descent, if not for the fact that the state $\mbf{x}$ does not belong to Euclidean space $\mathbb{R}^n$ but rather $SE(n)^{N-1}$. As such, the expression $ \p \mbf{g}(\mbf{x})/\p \mbf{x}$ is meaningless unless properly defined.

\subsection{On-manifold Cost and Gradient Descent}

The modification employed in this paper is to reparameterize the measurement model by defining $\mbf{x} = \mbfbar{x} \oplus \delta \mbf{x}$, leading to 
\beq 
\mbf{y} = \mbf{g}(\mbfbar{x} \oplus \delta \mbf{x}) + \mbf{v} \triangleq \mbfbar{g}(\delta \mbf{x}).
\eeq
The state $\mbfbar{x}$ will represent the current optimization iterate, which will be updated using $\delta \mbf{x}$.

Since the argument of the new measurement model $\mbfbar{g}(\delta \mbf{x})$ now belongs to Euclidean space $\mathbb{R}^{m(N-1)}$, it is possible to compute the ``local'' approximation to the FIM \cite{Bonnabel2015} at $\mbf{x} = \mbfbar{x}$ with $\mbf{I}(\mbfbar{x}) = \mbf{H}(\mbfbar{x})^\trans \mbf{R}^{-1} \mbf{H}(\mbfbar{x})$ where 
\bdis 
\mbf{H}(\mbfbar{x}) = \left.\frac{\p \mbf{g}(\mbfbar{x} \oplus \delta \mbf{x})}{\p \delta \mbf{x}}\right|_{\delta \mbf{x} = \mbf{0}},
\edis 
and evaluate the cost function $J_{\mathrm{est}} (\mbfbar{x}) = - \ln \det \mbf{I}( \mbfbar{x})$. Finally, an on-manifold gradient descent step can be taken with
\beq 
\mbfbar{x} \gets \mbfbar{x} \oplus\left( -\gamma \left.\frac{\p J_{\mathrm{est}} (\mbfbar{x} \oplus \delta \mbf{x})}{\p \delta \mbf{x}}\right|_{\delta\mbf{x} = \mbf{0}} \right)^\trans,
\eeq
where $\gamma$ is a step size.

The proposed gradient descent procedure is actually a standard approach to optimization on matrix manifolds \cite{Absil2008}. From a differential-geometric point of view, an approximation to the FIM is computed in the tangent space of the current optimization iterate $\mbfbar{x}$, which is a familiar Euclidean vector space. A gradient descent step is computed in the tangent space, and the result is \emph{retracted} back to the manifold $SE(n)^{N-1}$ using the retraction $\mc{R}_{\mbfbar{x}}(\delta \mbf{x}) = \mbfbar{x} \oplus \delta \mbf{x}$.


\begin{figure}
    \centering
    \includegraphics[width =0.5\linewidth, clip=true, trim={1cm 0cm 2cm 0cm}]{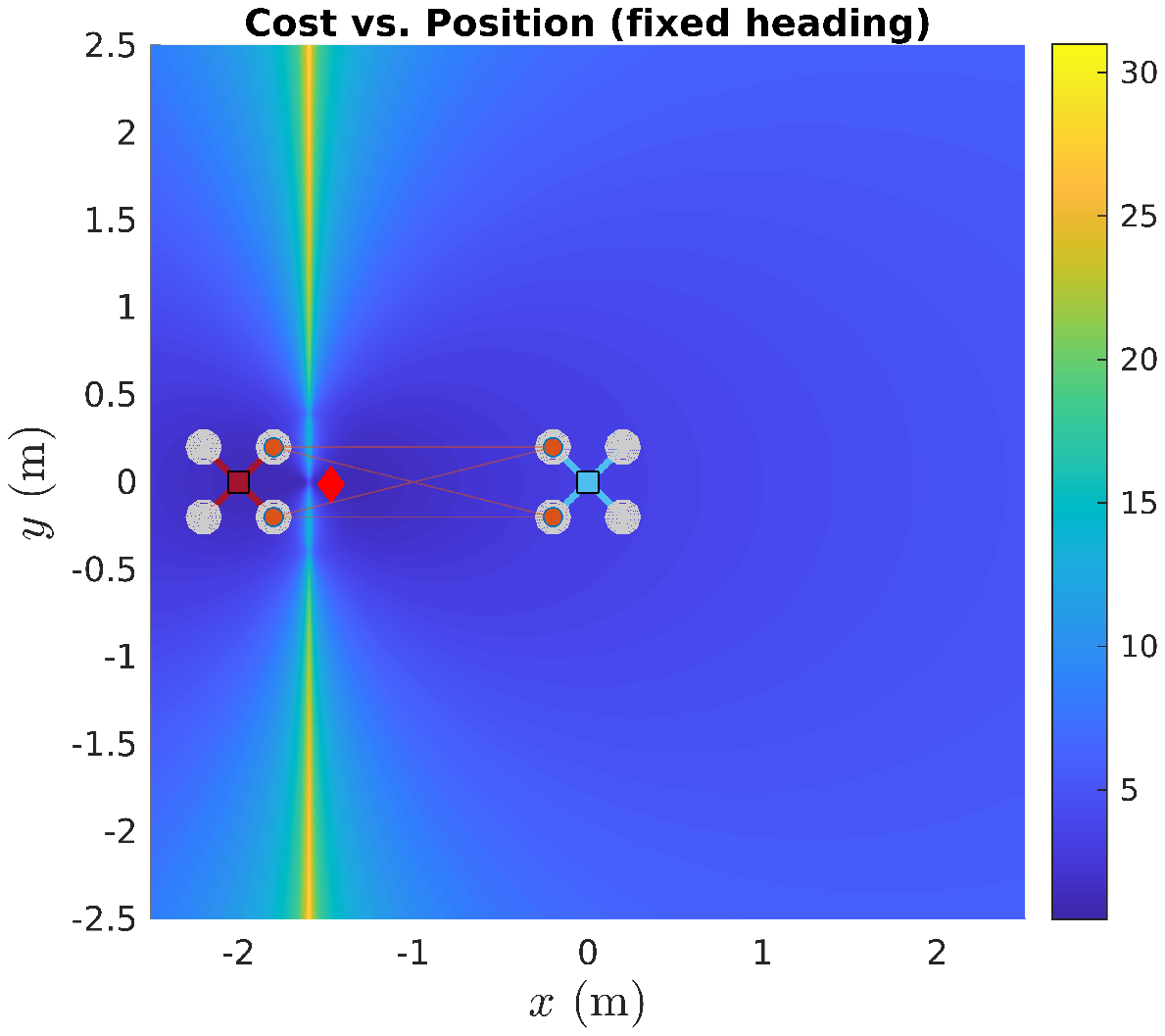}%
    \includegraphics[width =0.5\linewidth, clip=true, trim={1cm 0cm 2cm 0cm}]{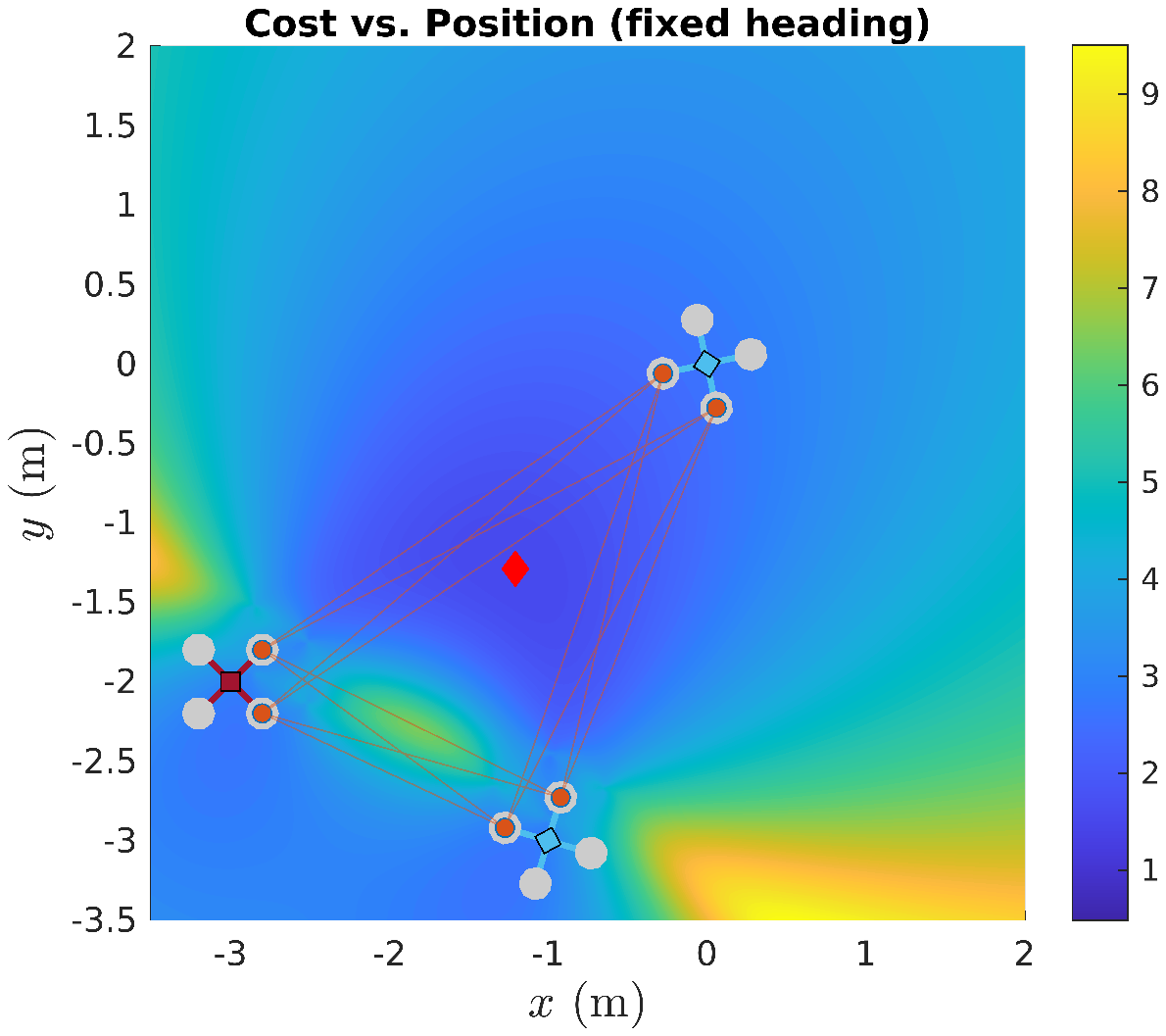}
    \includegraphics[width =0.5\linewidth, clip=true, trim={1cm 0cm 2cm 0cm}]{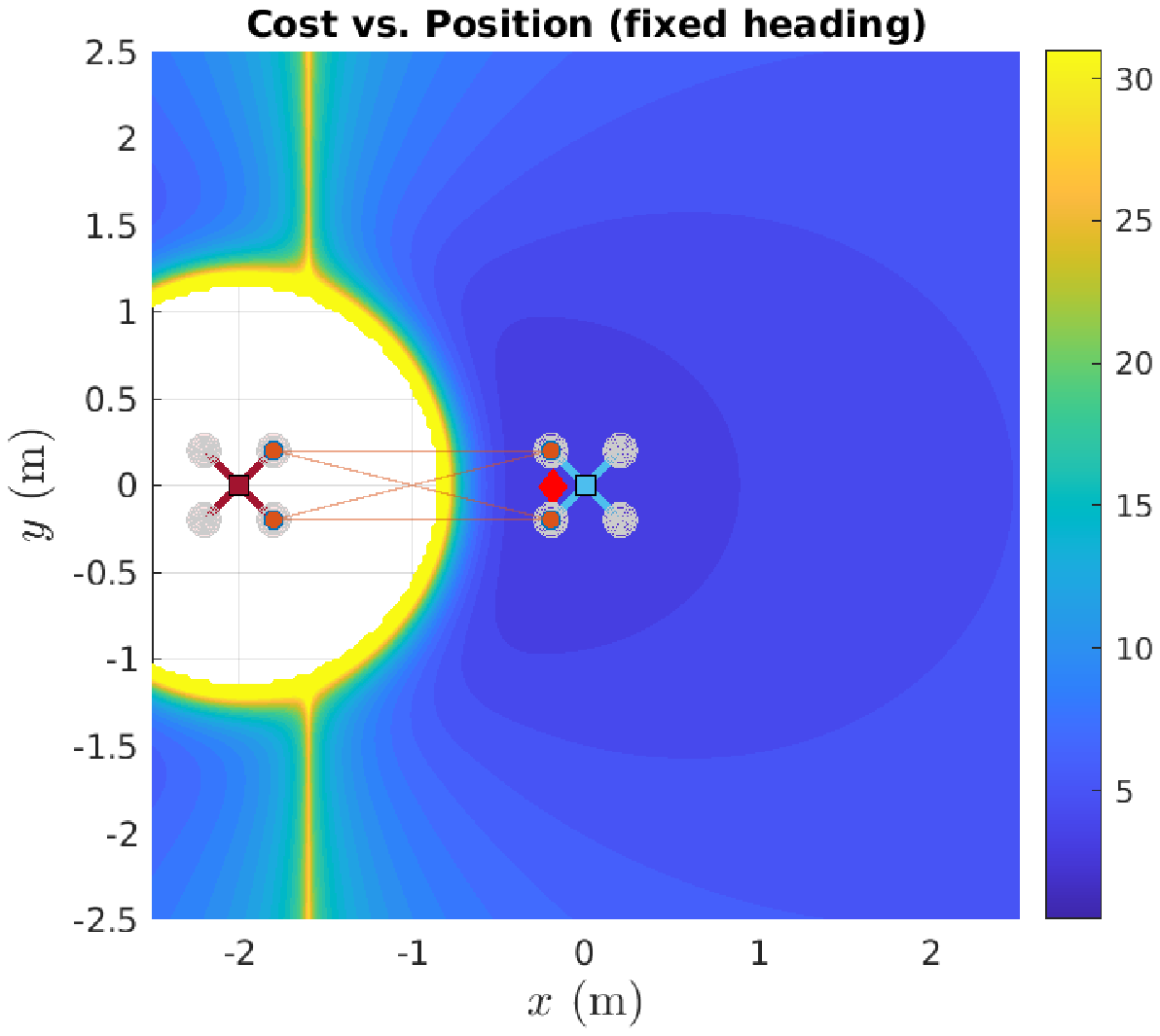}%
    \includegraphics[width =0.5\linewidth, clip=true, trim={1cm 0cm 2cm 0cm}]{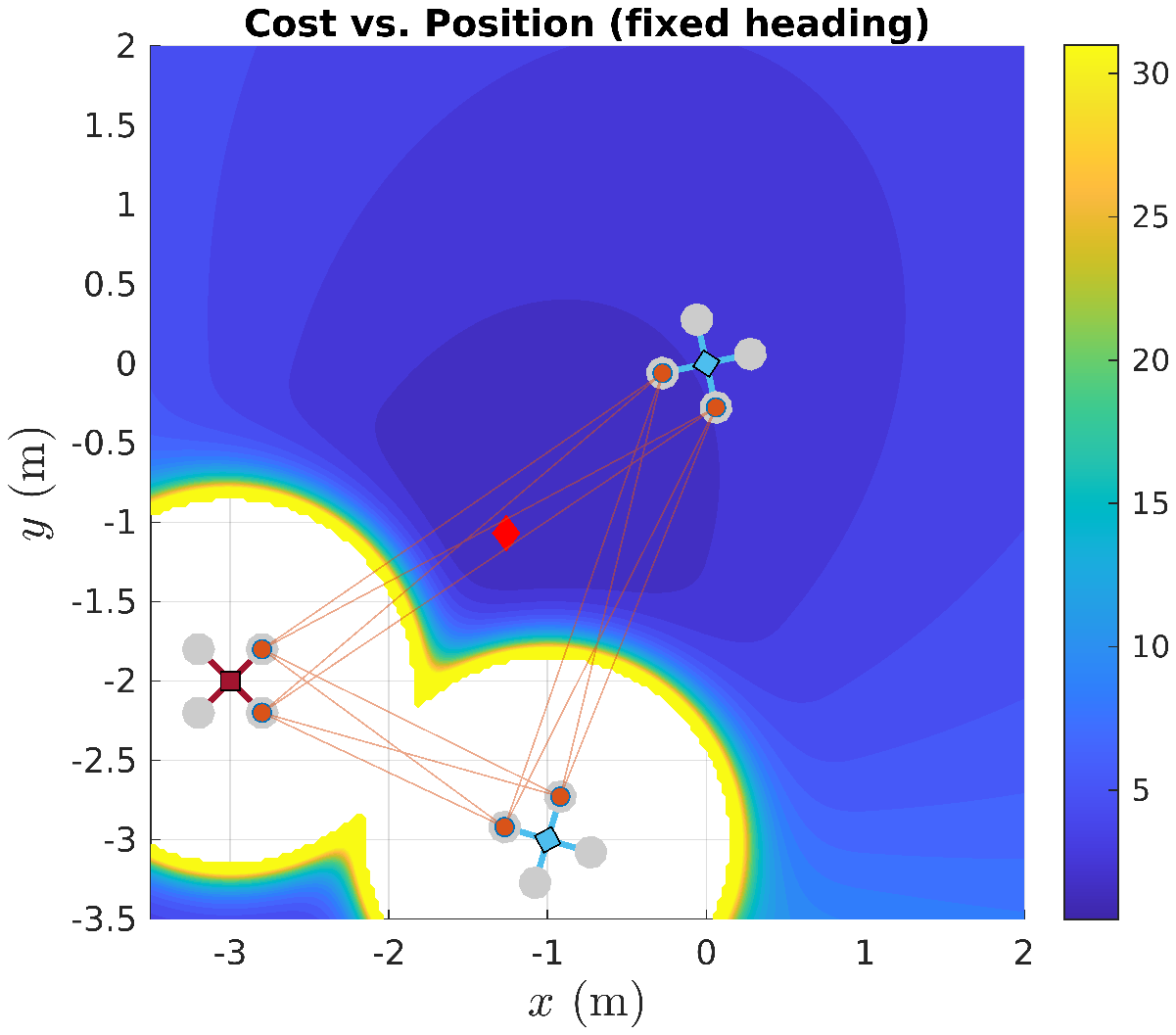}
    \caption{All four plots show the value of the cost with varying agent position (right agent for two-agent scenario, top agent for three-agent scenario), while maintaining fixed heading. The top row shows only the estmation cost $J_\mathrm{est}$, while the bottom row shows the total cost $J$ including the collision avoidance term. }
    \label{fig:cost}
\end{figure}

\subsection{Cost function implementation} \label{sec:cost}
Creating an implementable expression for the cost function $J_\mathrm{est}( \mbfbar{x}) = -\ln \det\left(\mbf{H}( \mbfbar{x})^\trans \mbf{R}^{-1} \mbf{H}( \mbfbar{x})\right)$ eventually amounts to computing the Jacobian of the range measurement model \eqref{eq:meas_model} with respect to $\delta \mbs{\xi}_\alpha$ and $\delta \mbs{\xi}_\beta$. To see this, 

\bdis 
\mbf{H}(\mbfbar{x}) = \bma{c} \vdots \\ \mbf{H}^{ij}( \mbfbar{x}) \\ \vdots \ema, 
\edis
\bdis
\mbf{H}^{ij}( \mbfbar{x}) = [\mbf{0}\ldots \mbf{H}^{ij}_\alpha(\mbfbar{x}) \ldots \mbf{H}^{ij}_\beta(\mbfbar{x}) \ldots \mbf{0}],
\edis
where
\begin{align*}
    \mbf{H}^{ij}_\alpha(\mbfbar{x})  &\triangleq \left. \frac{\p y_{ij}(\mbfbar{x}\oplus \delta \mbf{x})}{\p \delta \mbs{\xi}_\alpha}\right|_{\delta \mbf{x} = \mbf{0}},  \\
    \mbf{H}^{ij}_\beta (\mbfbar{x}) &\triangleq \left.\frac{\p y_{ij}(\mbfbar{x}\oplus \delta \mbf{x})}{\p \delta \mbs{\xi}_\beta}\right|_{\delta \mbf{x} = \mbf{0}}.
\end{align*}
The row matrix $\mbf{H}^{ij}( \mbfbar{x}) \in \mathbb{R}^{1 \times m(N-1)}$ represents the Jacobian of a single range measurement $y_{ij}$ with respect to the full state perturbation $\delta \mbf{x}$. This resulting matrix will be zero everywhere except for two blocks $\mbf{H}^{ij}_\alpha( \mbfbar{x})  \in \mathbb{R}^{1 \times m}$ and $\mbf{H}^{ij}_\beta( \mbfbar{x})  \in \mathbb{R}^{1 \times m}$, respectively located at the $\alpha^{\textrm{th}}$ and $\beta^{\textrm{th}}$ block columns, and have closed-form expressions derived in Appendix \ref{sec:meas_jac}. 

The cost function $J_\mathrm{est}$ is visualized for varying agent position in the top row of Figure \ref{fig:cost}, where the red dot shows the minimum found within that view. Looking at the top-left plot of Figure \ref{fig:cost}, there is a vertical line of high cost near the agent on the left, corresponding exactly to when all four tags line up, leading to an unobservable formation. Similarly, the three-agent scenario in the top-right plot of Figure \ref{fig:cost} shows a high cost when the agents are nearly all on the same line, which is a situation of near-unobservability. However, as can be seen in the top-left plot, the minimum is unacceptably close to the left agent, which would cause them to collide. Indeed, we have observed that naively descending the cost $J_\mathrm{est}$ alone leads to all the agents collapsing into each other. An explanation for this  behavior is that when agents are closer together, changes in attitude result in larger changes in the range measurements, which increases Fischer information. Nevertheless, in practice, collisions must be avoided, and this is done by augmenting the cost with an additional collision avoidance term $J_\mathrm{col}(\mbf{x})$, such that the total cost $J(\mbf{x})$ is 
\bdis 
J(\mbf{x}) = J_\mathrm{est}(\mbf{x}) +  J_\mathrm{col}(\mbf{x}), \qquad J_{\mathrm{col}}(\mbf{x}) = \sum_{\substack{\alpha,\beta \in \mc{A} \\ \alpha \neq \beta}} J_\mathrm{col}^{\alpha\beta}(\mbf{x}),
\edis
where a collision avoidance cost from \cite{Xia2016} is used,
\beq
J_\mathrm{col}^{\alpha\beta}(\mbf{x}) = \left(\min \left\lbrace 0 , \frac{\norm{\mbf{r}^{a_\alpha a_\beta}_1}^2 - R^2}{\norm{\mbf{r}^{a_\alpha a_\beta}_1}^2 - d^2} \right\rbrace \right)^2.
\eeq
The term $R$ represents an ``activation radius'' and $d$ is the safety collision avoidance radius. In this paper, the agent relative position is expressed as a function of pose matrices with 
\bdis
\mbf{r}^{a_\alpha a_\beta}_1 = \mbf{D}\mbf{T}_{1\alpha}\mbf{b} - \mbf{D}\mbf{T}_{1\beta} \mbf{b},
\edis
where $\mbf{D}=[\mbf{1} \; \mbf{0}],$ $\mbf{b} = [\mbf{0}\; 1]^\trans$. The new cost function $J$ is plotted on the bottom row of Figure \ref{fig:cost}, showing the effect of the collision avoidance term. Finally, one is now ready to descend the cost directly with 
\beq 
\mbfbar{x} \gets \mbfbar{x} \oplus\left( -\gamma \left.\frac{\p J (\mbfbar{x} \oplus \delta \mbf{x})}{\p \delta \mbf{x}}\right|_{\delta\mbf{x} = \mbf{0}} \right)^\trans.\label{eq:final_gd}
\eeq
In this work, the Jacobian of $J_\mathrm{est}$ is computed numerically with finite difference \cite{Cossette2020a}, and the optimization is only done offline for the following reasons. The solution to the optimization problem is only a function of some physical properties, the measurement graph $\mc{G}$, and the number of robots $N$. For any experiments that use the same hardware, the physical properties such as the safety radius, tag locations, and measurement covariances, all remain constant. The measurement graph $\mc{G}$ can often also be assumed to be constant and fully connected. Even though full-connectedness is not necessary to find optimal formations using the proposed approach, technologies such as UWB often have a ranging limit that is well beyond the true ranges between all robots in the experiment. Hence, it is straightforward to precompute optimal formations for varying robot numbers $N$ with fully-connected measurement graphs, and to store the solutions in memory onboard each robot.

Nevertheless, a distributed, real-time implementation is required for varying measurement graphs, which is likely to arise in the presence of obstacles that block line-of-sight. Such a scenario requires simultaneously satisfying obstacle avoidance constraints and perhaps other planning objectives, which is beyond the scope of this paper.

\subsection{Optimization results}
\begin{figure}[t]
    \centering
    \includegraphics[width =0.5\linewidth, clip=true, trim={2cm 0cm 2cm 0cm}]{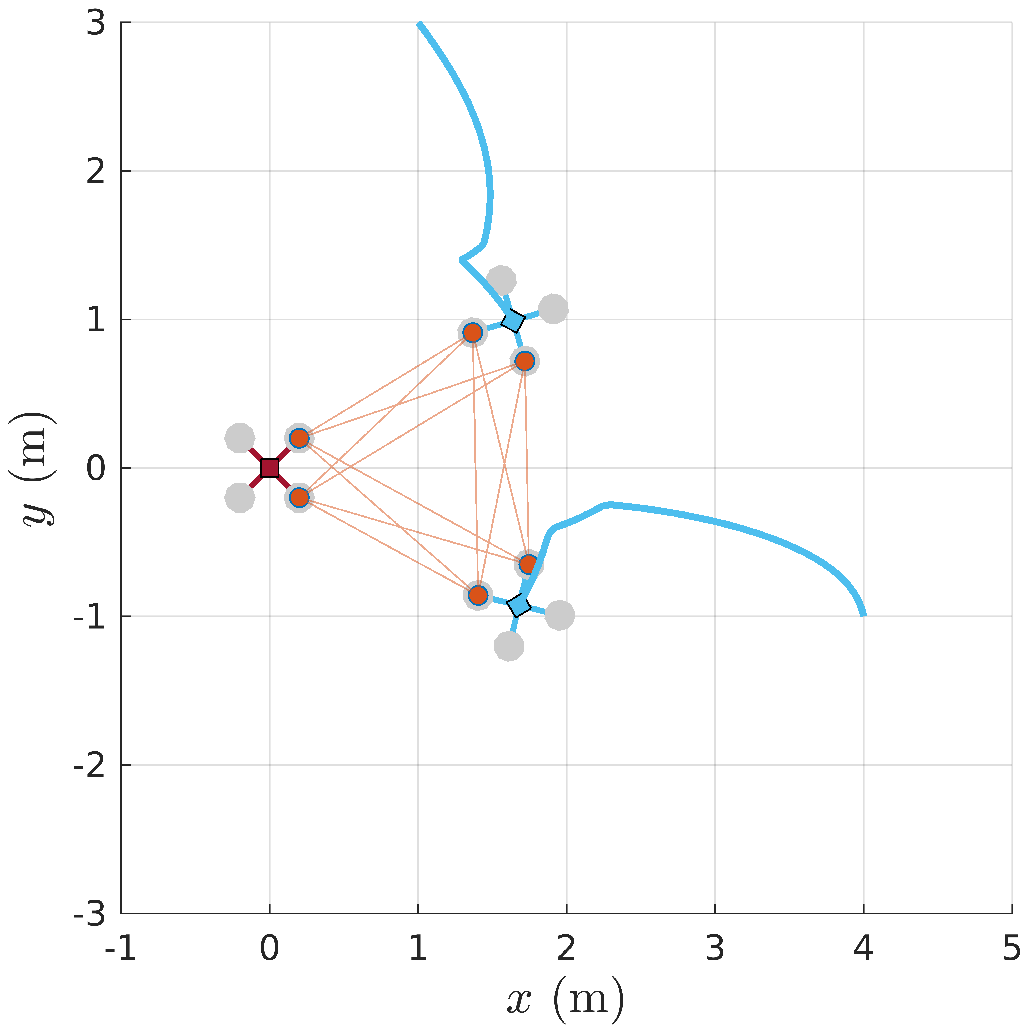}%
    \includegraphics[width =0.5\linewidth, clip=true, trim={2cm 0cm 2cm 0cm}]{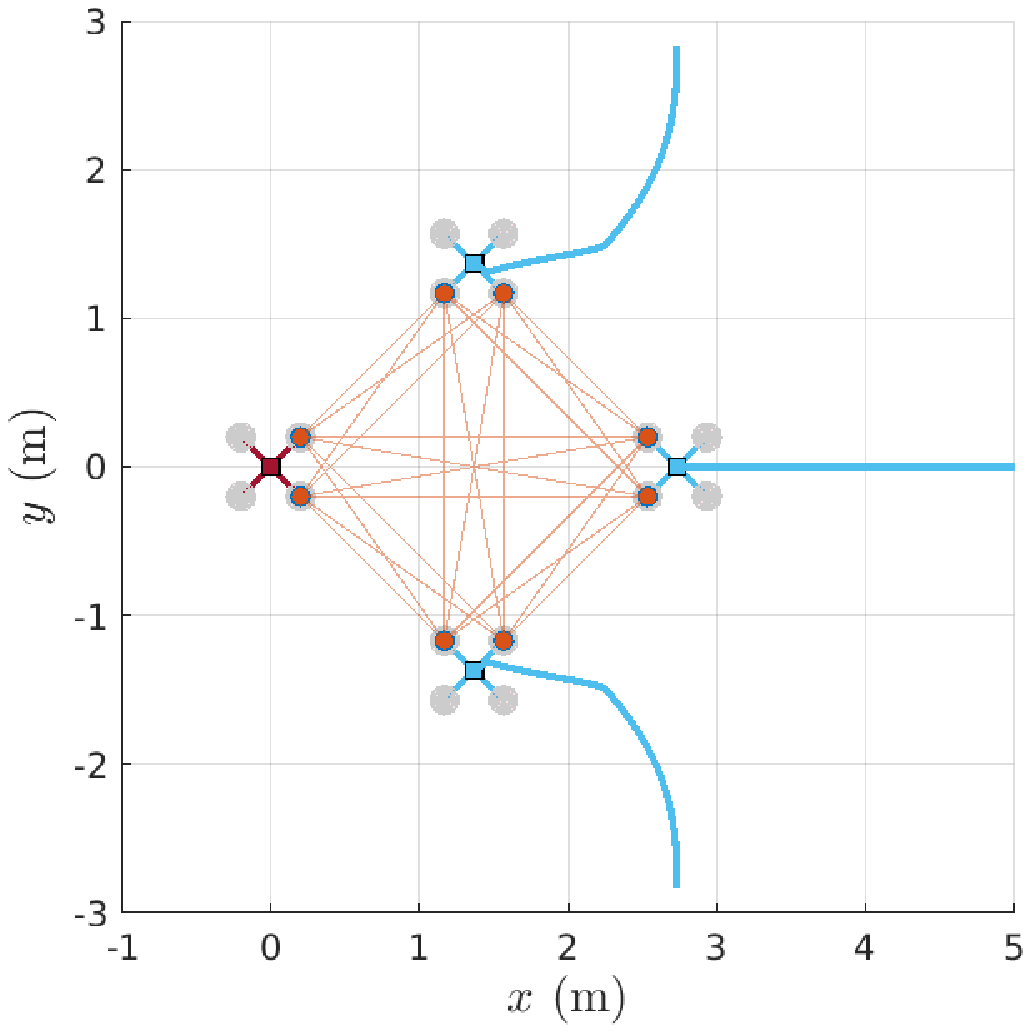}
    \includegraphics[width =0.5\linewidth, clip=true, trim={2cm 0cm 2cm 0cm}]{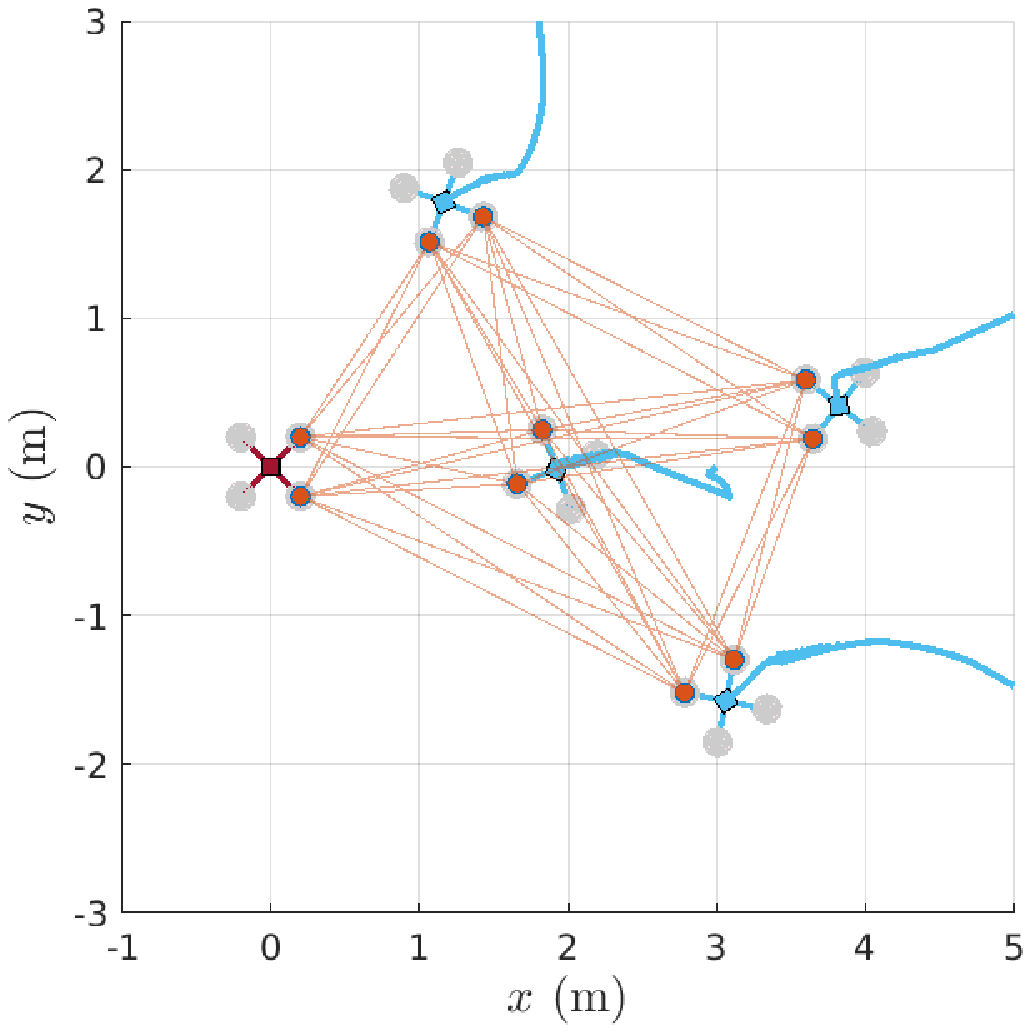}%
    \includegraphics[width =0.5\linewidth, clip=true, trim={2cm 0cm 2cm 0.8cm}]{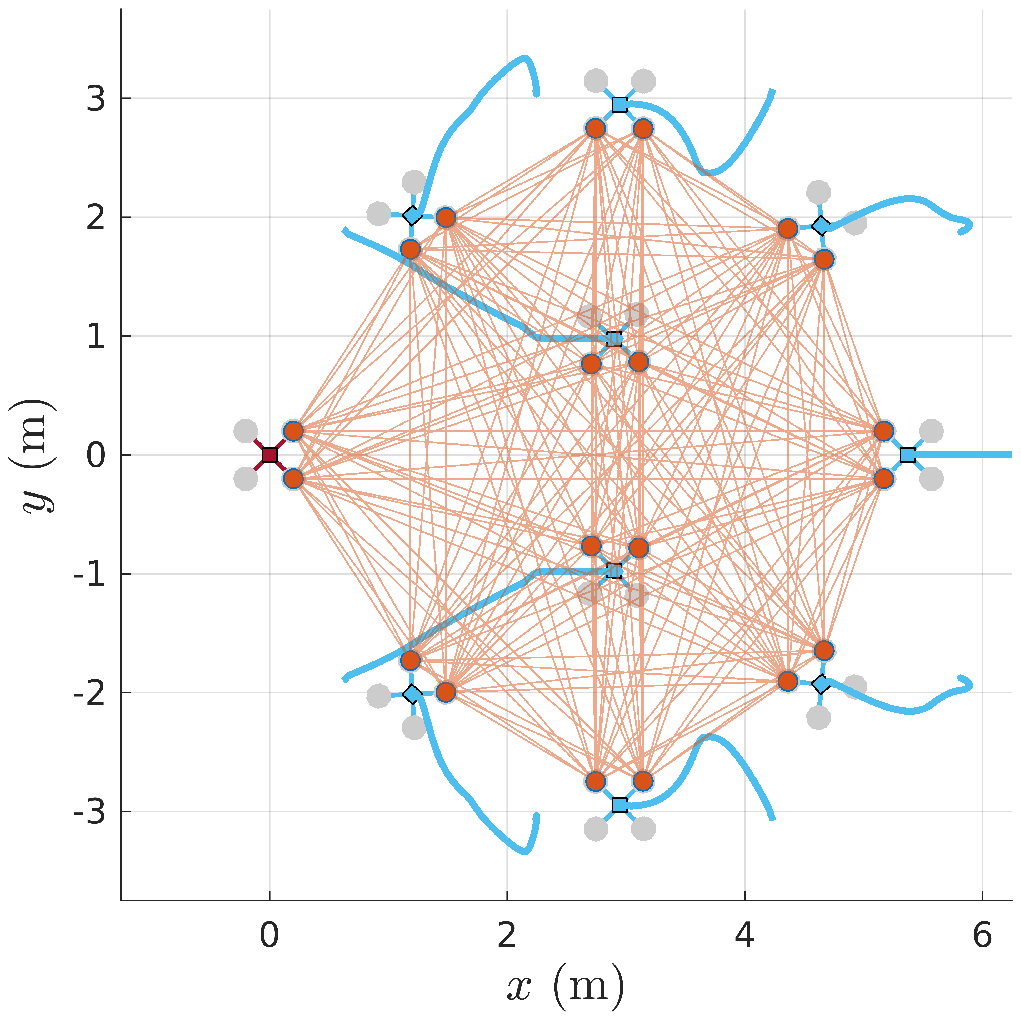}
    \caption{Final locally optimal formations for 3, 4, 5, and 10 agents.}
    \label{fig:opt1}
\end{figure}

The gradient descent in \eqref{eq:final_gd} is performed with a step size of $\gamma = 0.1$, an activation radius of $R=2~\mathrm{m}$, and a safety radius of $d = 1~\mathrm{m}$. Each agent has two tags located at 
\bdis 
\mbf{r}^{\tau_i a_\alpha}_\alpha = [0.2\; 0.2]^\trans, \quad \text{and} \quad \mbf{r}^{\tau_j a_\alpha}_\alpha = [0.2\; -0.2]^\trans,
\edis
where $\alpha, i, j$ are arbitrary and the units are in meters. Figure \ref{fig:opt1} shows quadcopter formations at convergence for 3, 4, 5, and 10 agents, each with a fully-connected measurement graph $\mc{G}$, except for edges corresponding to two tags on the same agent. The results shown here are intuitive, with the three- and four-agent scenarios corresponding to an equilateral triangle and square, respectively. However, with increasing agent numbers, regular polygon formations are no longer optimal, as can be seen in the five- and ten-agent scenarios.

\begin{figure}[t]
    \centering
    \includegraphics[width =0.5\linewidth, clip=true, trim={1cm 0cm 2cm 0cm}]{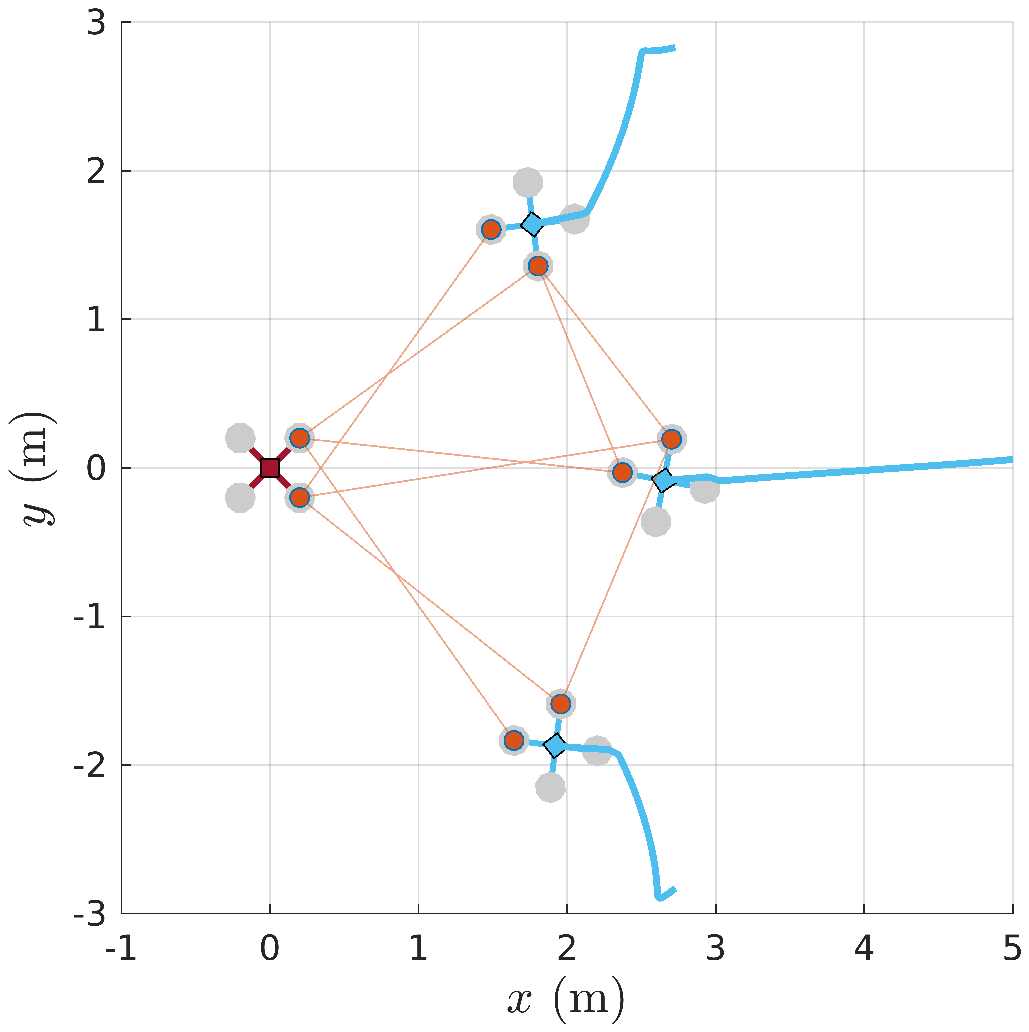}%
    \includegraphics[width =0.5\linewidth, clip=true, trim={1cm 0cm 2cm 0cm}]{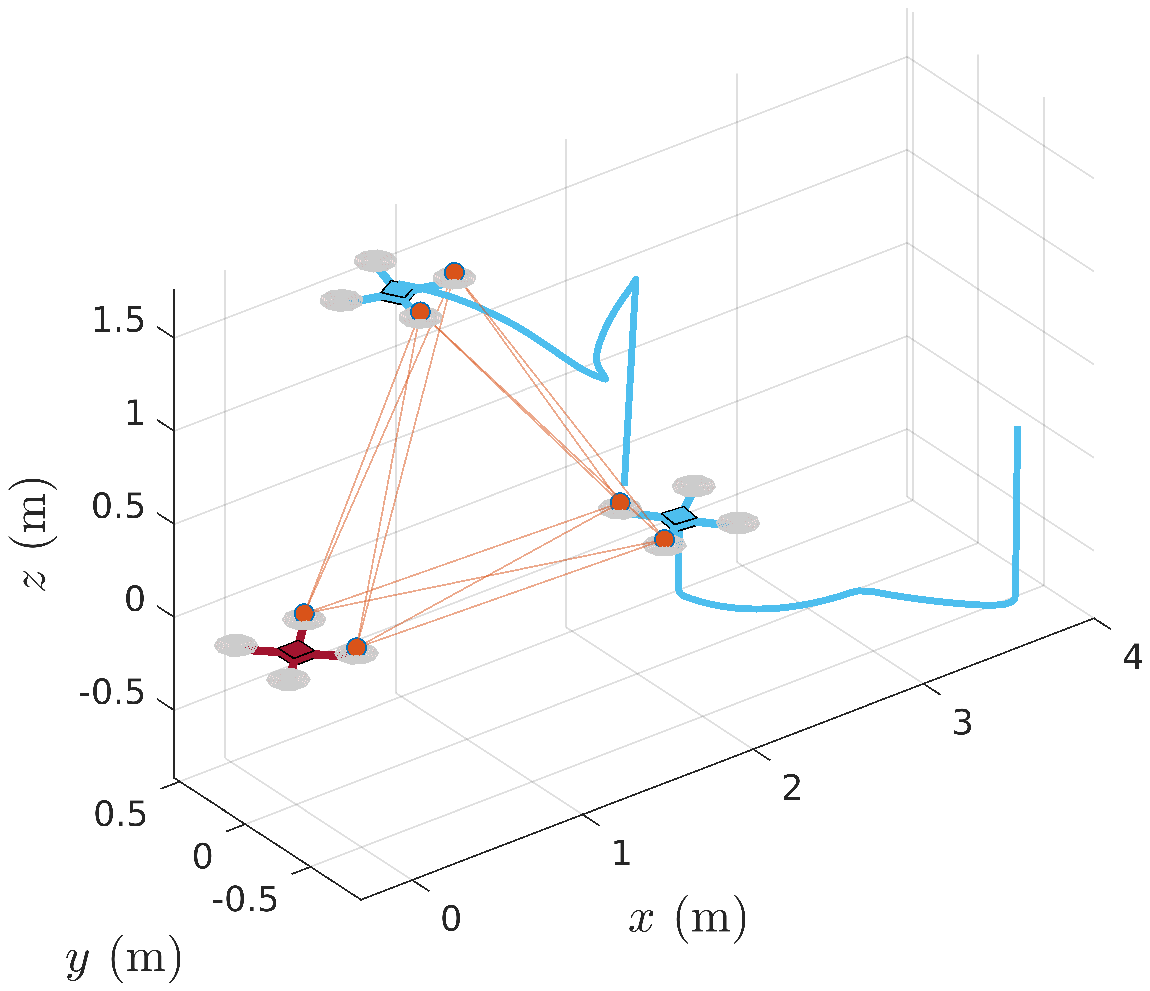}
    \caption{(left) Optimal formation with sparse measurement graph. (right) Optimal formation with 3D position and heading as design variables.}
    \label{fig:opt2}
\end{figure}

Since the treatment in this paper is general to an arbitary measurement graph $\mc{G}$, provided the FIM remains maximum rank, optimization is also performed for a non-fully-connected measurement graph. The results for this along with a 3D scenario are shown in Figure \ref{fig:opt2}. In 3D, the robot relative poses are represented with elements of $SE(3)$. However, since the presented simulations contain only two-tag robots, relative roll and pitch between robots are unobservable, which would make the cost infinite, unless more sensors are used. Hence, roll and pitch are excluded from the optimization and their values are fixed to zero. This leaves the three translational components and heading as the four degrees of freedom available for optimization. This is easily implemented in practice with a redefinition of the ``wedge'' operator such that $(\cdot)^\wedge:\mathbb{R}^4 \to \mathfrak{se}(3)$. Moreover, from an application  standpoint, both ground vehicles and quadcopter-type aerial vehicles only have heading as a rotational degree of freedom available for planning.


\subsection{Validation on a least squares estimator} \label{sec:eval}
To validate the claim that descending the cost improves the estimation performance, a non-linear least-squares estimator is used. At regular iterates $\mbfbar{x}$ of the optimization trajectory, a small 2000-trial Monte Carlo experiment is performed, where in each trial a set of range measurements are generated with $\mbf{y} = \mbf{g}(\mbfbar{x}) + \mbf{v},$ $\mbf{v} = \mc{N}(\mbf{0}, \mbf{R})$. Then, an on-manifold Gauss-Newton procedure \cite{Barfoot2019} is used to solve 
\beq 
\mbfhat{x} = \arg \min_{\mbf{x}} \frac{1}{2} \sum_{\alpha=2}^N\norm{\ln(\mbf{C}_{1\alpha}^\trans \mbfcheck{C}_{1\alpha})^\vee}^2_{\mbfcheck{P}_\alpha} + \frac{1}{2}\norm{\mbf{y} - \mbf{g}(\mbf{x})}^2_{\mbf{R}}, \label{eq:estimator}
\eeq
where $\norm{\mbf{e}}_\mbf{M}^2 = \mbf{e}^\trans \mbf{M}^{-1} \mbf{e}$ denotes a squared Mahalanobis distance, and an attitude prior with ``mean'' $\mbfcheck{C}_{1\alpha}$  and
covariance $\mbfcheck{P}_{\alpha}$ is also included for each agent. It turns out that minimization of only the second term in \eqref{eq:estimator} yields unacceptably poor estimation performance, as the solution often converges to local minimums depending on
the initial guess. The inclusion of an attitude prior, which is practically obtained by dead-reckoning on-board gyroscope measurements, yields much lower overall estimation error.

\begin{figure}[t]
    \centering
    \includegraphics[width = \linewidth,clip, trim={5cm 4.8cm 5cm 5cm}]{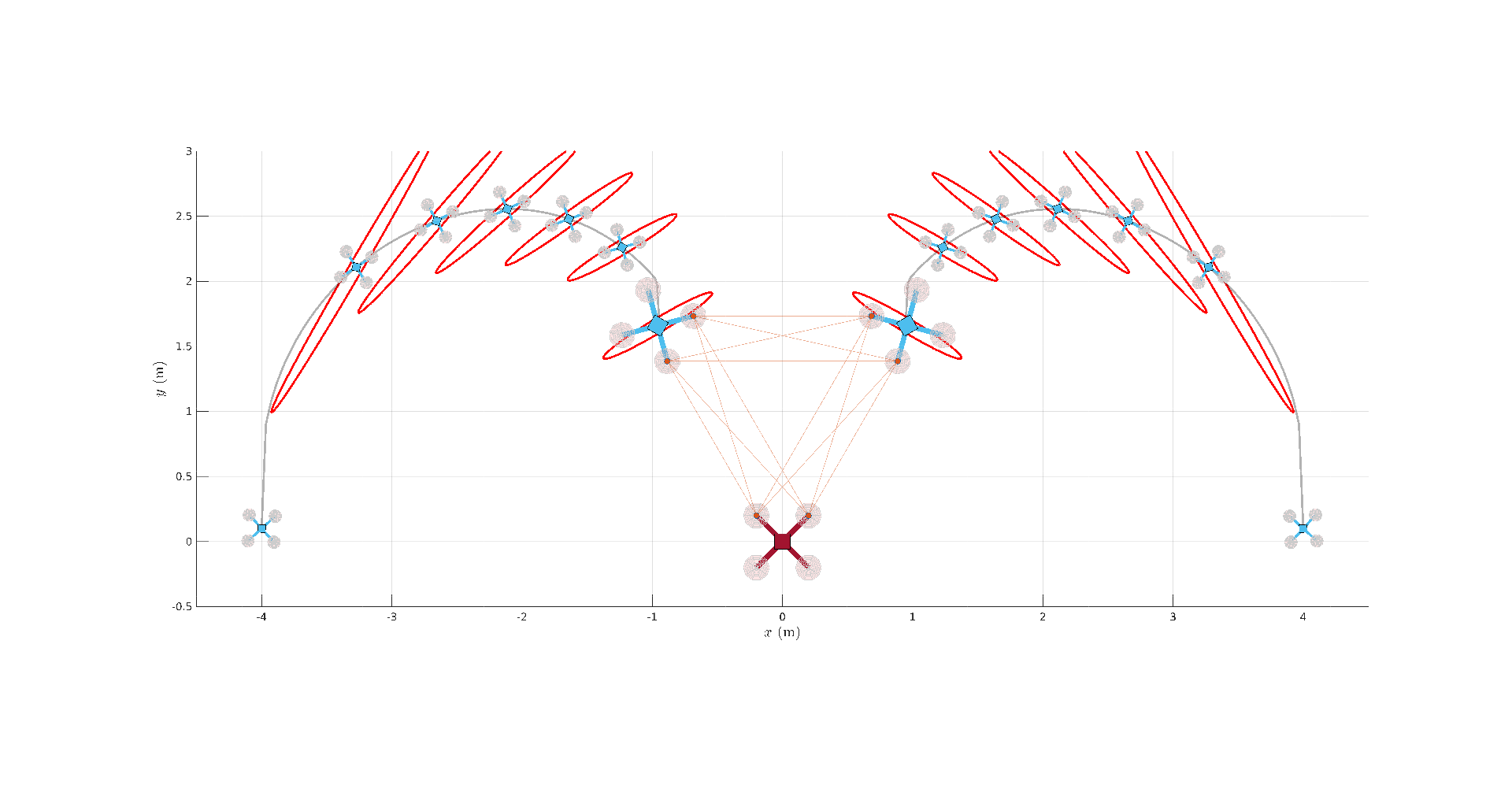}
    \caption{Trajectory taken during optimization with superimposed $1\sigma$ equal-probability contours corresponding to the Cram\'er-Rao bound. The ellipsoids for the starting positions are too large to fit in the figure.}
    \label{fig:snapshot}
\end{figure}

\begin{figure}
    \centering
    \includegraphics[width = \linewidth]{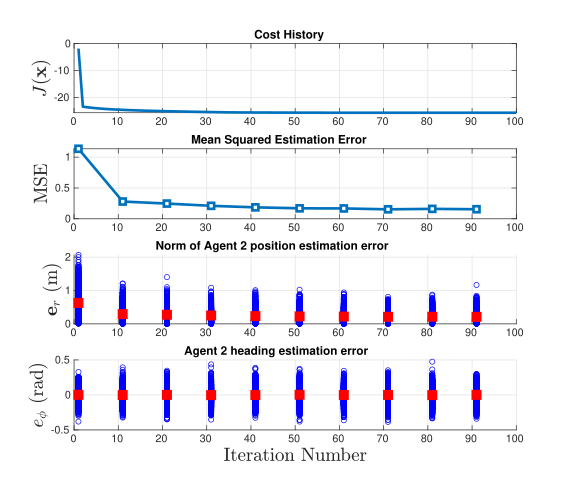}
    \caption{Cost function, along with various metrics of a least-squares estimator, obtained from 2000 Monte-Carlo trials at various points during the optimization. In the bottom two plots, the red squares denote the average norm of the respective estimation errors.}
    \label{fig:least_squares}
\end{figure}

Figure \ref{fig:least_squares} shows the value of the cost throughout the optimization trajectory, as well as the mean squared estimation error over the $K = 2000$ Monte Carlo trials per optimization step. The true agent poses are initialized in a near-straight line, as shown in Figure \ref{fig:snapshot}, and the covariances used are $\mbf{R}= 0.1^2 \mbf{1}~\mathrm{m}^2$, $\mbfcheck{P}_\alpha = 0.08^2~\mathrm{rad}^2$. The mean squared estimation error (MSE) is calculated with 
\beq
\mathrm{MSE} = \frac{1}{K}\sum_{k = 1}^{K} \delta \mbs{\xi}^\trans \delta \mbs{\xi}, \qquad \delta \mbs{\xi} = \bma{c} \ln(\mbfbar{T}_{12}^{-1}\mbfhat{T}_{12})^\vee \\ \vdots \\ \ln(\mbfbar{T}_{1N}^{-1}\mbfhat{T}_{1N})^\vee \ema,
\eeq
and shows a clear correlation with the cost function. This provides evidence for the fact that descending the proposed cost function also reduces the estimation error.

\section{Experimental Evaluation} \label{sec:exp}
\begin{figure}
    \centering
    \includegraphics[width = \linewidth]{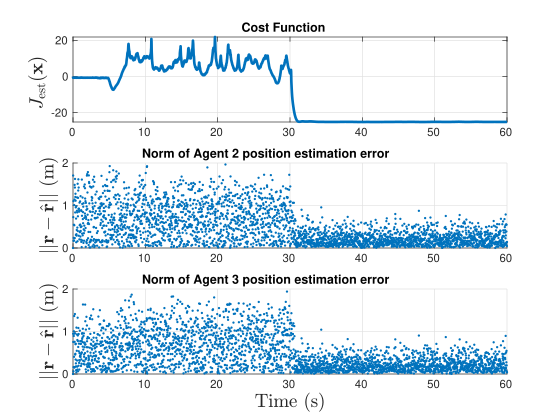}
    \caption{Experimental results using a least squares estimator. From 0 s to 30 s, the quadcopters are in a line formation and the average positioning error is 0.77 m. From 30 s to 60 s, the quadcopters are in an optimal formation and the average positioning error is 0.22 m, a 68\% reduction.}
    \label{fig:experiment}
\end{figure}

\begin{figure}
    \centering
    \includegraphics[width = 0.9\linewidth]{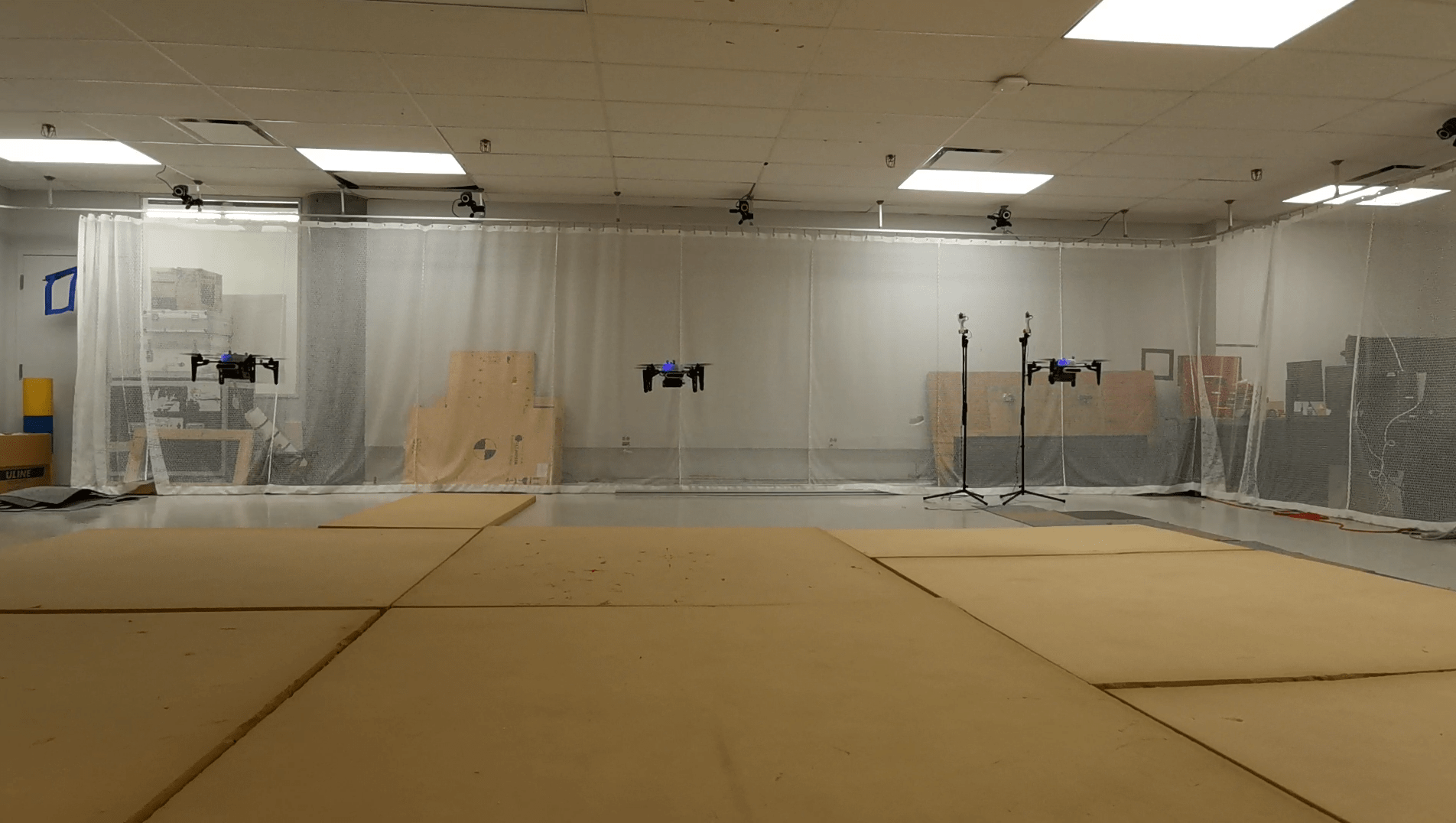}\\
    \includegraphics[width = 0.9\linewidth]{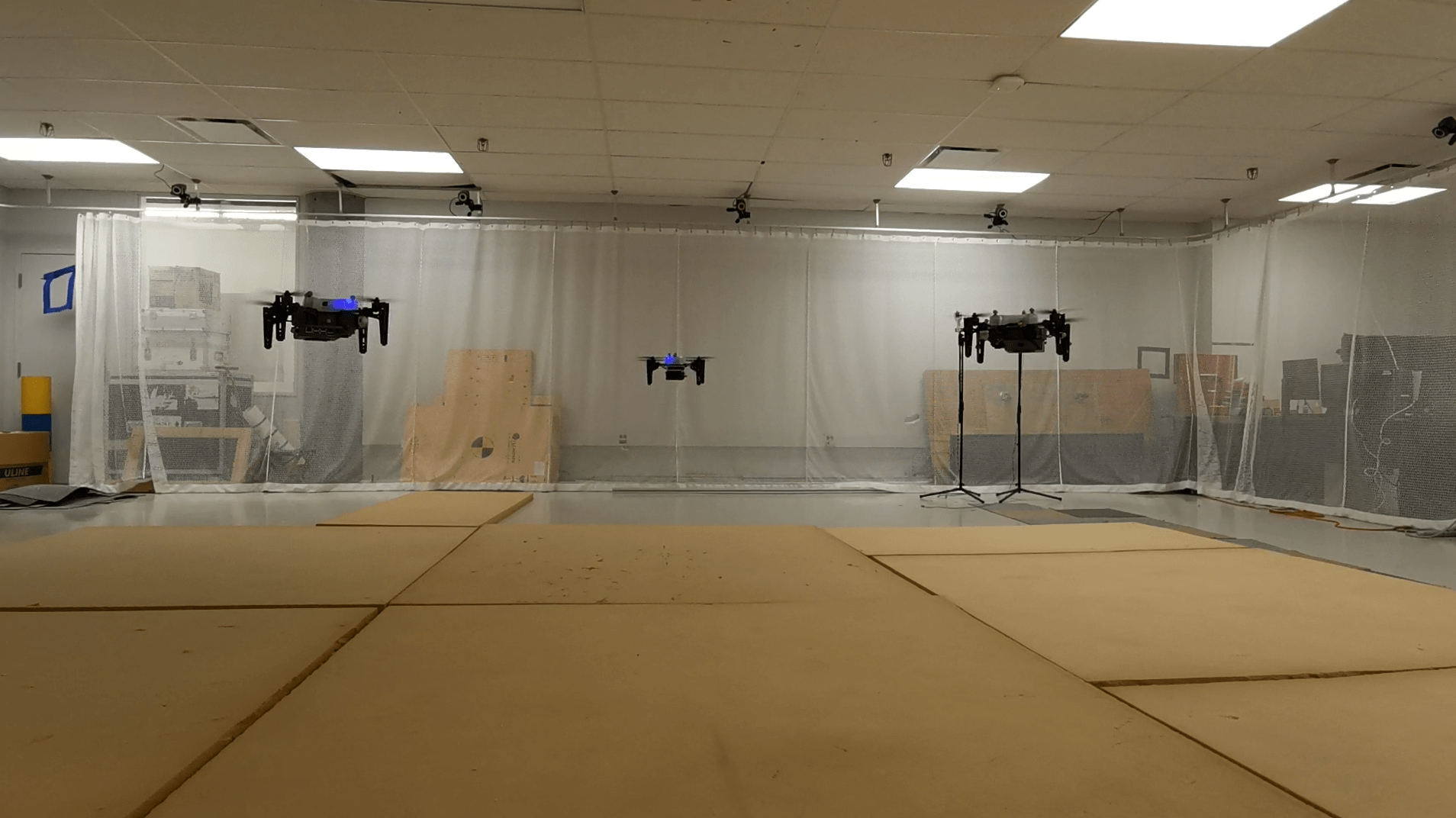}
    \caption{(top) Three quadcopters in an initial straight line formation. (bottom) Quadcopters in an optimal triangle formation.
    }
    \label{fig:drones}
\end{figure}

An estimator is also run with three PX4-based Uvify IFO-S quadcopters in order to experimentally validate the claim that descending the proposed cost function results in improved estimation performance. The quadcopters start by flying in a line formation and, after 30 seconds, proceed to a triangle formation computed using the proposed framework for another 30 seconds, as shown in Figure \ref{fig:drones}. Figure \ref{fig:experiment} shows the position estimation error using the least-squares estimator presented in Section \ref{sec:eval}. Real gyroscope measurements are used to obtain an attitude prior at all times. Range measurements are synthesized with a standard deviation of 10 cm using ground truth vehicle poses obtained from a motion capture system. The UWB tags are simulated to be 17 cm apart, corresponding to extremities of the propeller arms. As can be seen in Figure \ref{fig:experiment}, moving to the optimal triangle formation, from one of the worst starting formations results in a 68\% reduction in estimation variance.

\section{Conclusion}
This paper shows, in both simulation and experiment, that range-based relative state estimation performance can be substantially improved by a proper choice of formation geometry. The largest improvements are obtained when the robots move away from unobservable formations.

The generalizability of the cost function makes it appropriate for use beyond direct minimization. For instance, consider using this function to impose an inequality constraint on an application-oriented planning problem, such as indoor exploration. Using an inequality constraint would allow robots the freedom to move within the feasible region in order to accomplish tasks such as infrastructure inspection, yet still avoid the ``worst'' formations with very high cost, which could cause problematically large state estimation errors. Future work may tackle a scenario similar to this, including developing a distributed computation scheme for the proposed cost function.

\appendix
\subsection{Measurement model Jacobian} \label{sec:meas_jac}
Let $y_{ij}(\mbf{x}) = y_{ij}(\mbfbar{x}) + \delta y_{ij} \triangleq \bar{y}_{ij} + \delta y_{ij}$, $\mbf{T}_{1\alpha} = \mbfbar{T}_{1\alpha}\exp(\delta \mbs{\xi}_\alpha^\wedge)$, $\mbf{T}_{1\beta} = \mbfbar{T}_{1\beta}\exp(\delta \mbs{\xi}_\beta^\wedge)$, and $v_{ij} = 0$. The terms $\delta y_{ij}, \delta \mbs{\xi}_\alpha, \delta \mbs{\xi}_\beta$ are assumed to be small quantities, which motivates, for example, the approximation $\exp(\delta \mbs{\xi}_\alpha^\wedge) \approx \mbf{1} + \delta \mbs{\xi}_\alpha^\wedge$. Equation \eqref{eq:meas_model} becomes

\begin{multline*}
    (\bar{y}_{ij} + \delta y_{ij})^2 = \left(\mbf{D}\mbfbar{T}_{1\alpha}(\mbf{1} + \delta \mbs{\xi}_\alpha^\wedge) \mbf{p}_i - \mbf{D}\mbfbar{T}_{1\beta}(\mbf{1} + \delta \mbs{\xi}_\beta^\wedge) \mbf{p}_j \right)^\trans\\
    \left(\mbf{D}\mbfbar{T}_{1\alpha}(\mbf{1} + \delta \mbs{\xi}_\alpha^\wedge) \mbf{p}_i - \mbf{D}\mbfbar{T}_{1\beta}(\mbf{1} + \delta \mbs{\xi}_\beta^\wedge) \mbf{p}_j \right),
\end{multline*}
which, after expanding and neglecting higher-order terms, leads to 
\begin{multline}
    2\bar{y}_{ij} \delta y_{ij} = 2 (\mbf{p}_i^\trans \mbfbar{T}_{1\alpha}^\trans\mbf{D}^\trans  - \mbf{p}_j^\trans \mbfbar{T}_{1\beta}^\trans \mbf{D}^\trans) \mbf{D}\mbfbar{T}_{1 \alpha} \delta \mbs{\xi}^\wedge_\alpha \mbf{p}_i \\
    +  2 (\mbf{p}_j^\trans \mbfbar{T}_{1\beta}^\trans\mbf{D}^\trans  - \mbf{p}_i^\trans \mbfbar{T}_{1\alpha}^\trans\mbf{D}^\trans ) \mbf{D}\mbfbar{T}_{1 \beta} \delta \mbs{\xi}^\wedge_\beta \mbf{p}_i. \label{eq:jac1}
\end{multline}
Next, it is straightforward to define a simple operator $(\cdot)^\odot$, as per \cite{Barfoot2019}, such that $\mbs{\xi}^\wedge \mbf{p} = \mbf{p}^\odot \mbs{\xi}$. Rearranging \eqref{eq:jac1} yields
\begin{multline}
    \delta y_{ij} = \frac{(\mbf{p}_i^\trans \mbfbar{T}_{1\alpha}^\trans\mbf{D}^\trans  - \mbf{p}_j^\trans \mbfbar{T}_{1\beta}^\trans \mbf{D}^\trans)}{\bar{y}_{ij}} \mbf{D}\mbfbar{T}_{1 \alpha} \mbf{p}_i^\odot \delta \mbs{\xi}_\alpha \\
    -\underbrace{\frac{(\mbf{p}_i^\trans \mbfbar{T}_{1\alpha}^\trans\mbf{D}^\trans  - \mbf{p}_j^\trans \mbfbar{T}_{1\beta}^\trans \mbf{D}^\trans)}{\bar{y}_{ij}}}_{\triangleq \mbs{\rho}_{ij}} \mbf{D}\mbfbar{T}_{1 \beta} \mbf{p}_j^\odot \delta \mbs{\xi}_\beta. \label{eq:jac2}
\end{multline}
The term $\mbs{\rho}_{ij}$ is the physical unit direction vector between tags $i$ and $j$, resolved in Agent 1's body frame. From \eqref{eq:jac2} it then follows that 
\bdis
\frac{\p y_{ij}}{\p \delta \mbs{\xi}_\alpha} = \mbs{\rho}_{ij} \mbf{D} \mbfbar{T}_{1\alpha} \mbf{p}_i^\odot, \qquad \frac{\p y_{ij}}{\p \delta \mbs{\xi}_\beta} = -\mbs{\rho}_{ij} \mbf{D} \mbfbar{T}_{1\beta} \mbf{p}_j^\odot.
\edis

{\AtNextBibliography{\small}
\printbibliography}

\end{document}

%% file: Commands.tex
\newcommand{\ignore}[1]{}

\newcommand{\norm}[1]{\left\Vert#1\right\Vert} 
\newcommand{\mc}[1]{\mathcal{#1}}

\newcommand{\bma}[1]{\left[\begin{array}{ #1}}
\newcommand{\ema}{\end{array}\right]}

\DeclareMathAlphabet{\mbf}{OT1}{ptm}{b}{n}
\newcommand{\mbs}[1]{{\boldsymbol{#1}}}

\newcommand{\mbfdot}[1]{{\dot{\mbf{#1}}}}
\newcommand{\mbfbar}[1]{{\bar{\mbf{#1}}}}
\newcommand{\mbfhat}[1]{{\hat{\mbf{#1}}}}
\newcommand{\mbfcheck}[1]{{\check{\mbf{#1}}}}

\def\fdota{{\raisebox{-2pt}{\LARGE $\cdot$}}}
\def\fdotb{{\raisebox{-0.6ex}{ \kern0.2ex\raisebox{0.8ex}{\tiny $\hspace*{-1ex}\circ$}}}}
\def\fddota{{\raisebox{-2pt}{\LARGE $\cdot\hspace*{-0.2ex}\cdot$}}}
\def\fddotb{{\raisebox{-0.6ex}{ \kern0.2ex\raisebox{0.8ex}{\tiny $\hspace*{-1ex}\circ\circ$}}}}

\newtheorem{theorem}{Theorem} 
\newtheorem{definition}{Definition}

\newcommand{\p}{\partial}

\newcommand{\trans}{{\ensuremath{\mathsf{T}}}} 
\newcommand{\utimes}{ {\raisebox{-0.6ex}{ \kern-1.0ex\raisebox{0.6ex}{ \small $\mathsf{v}$}}} } %
\newcommand{\beq}{\begin{equation}}
\newcommand{\eeq}{\end{equation}}
\newcommand{\bdis}{\begin{displaymath}}
\newcommand{\edis}{\end{displaymath}}
\newcommand{\beqarray}{\begin{eqnarray}}
\newcommand{\eeqarray}{\end{eqnarray}}
\newcommand{\beqarraynn}{\begin{eqnarray*}}
\newcommand{\eeqarraynn}{\end{eqnarray*}}
\newcommand{\balign}{\begin{align}}
\newcommand{\ealign}{\end{align}}
\newcommand{\balignnn}{\begin{align*}}
\newcommand{\ealignnn}{\end{align}}

\renewcommand{\theenumii}{\arabic{enumii}}
\makeatletter
\renewcommand{\p@enumii}{\theenumi.}
\makeatother
